\documentclass[letterpaper]{article} 
\usepackage{aaai2027}  
\usepackage[hyphens]{url}  
\usepackage{graphicx} 
\usepackage{natbib}  
\usepackage{caption} 
\usepackage{algorithm}
\usepackage{algorithmic}

\usepackage{newfloat}
\usepackage{listings}
\DeclareCaptionStyle{ruled}{labelfont=normalfont,labelsep=colon,strut=off} 
\floatstyle{ruled}
\newfloat{listing}{tb}{lst}{}
\floatname{listing}{Listing}

\usepackage{booktabs}
\usepackage{tabularx}
\usepackage{array}
\newcolumntype{L}{>{\raggedright\arraybackslash}X}

\title{A Prompt-Engineering Approach to Develop Scalable, Flexible, and Real-Time Hybrid  Micro-Level Personalization in a General Purpose AI Teaching Assistant.}
\author{Saptarshi Basu, Sandeep Kakar, Ashok Goel}
\affiliations {
    Georgia Institute of Technology, Atlanta GA 30332, USA\\
    sbasu7@gatech.edu, skakar6@gatech.edu, ashok.goel@cc.gatech.edu
}

\begin{document}

\maketitle

\begin{abstract}

Artificial intelligence (AI) teaching assistants powered by large language models (LLMs) offer scalable educational support but often provide limited personalization. This study presents a prompt-engineering-based framework for personalizing general-purpose LLM/RAG based AI teaching assistant such as Jill Watson across academic disciplines and courses. The framework adapts responses using six learner-specific dimensions: self-assessment, abstraction preference, verbosity preference, perceptual orientation, information processing style, and level of understanding, yielding 96 distinct learner profiles. Student queries are additionally analyzed using Bloom’s Taxonomy to estimate cognitive complexity at the interaction level. Learner attributes and cognitive assessments are encoded in structured prompts that condition the LLM without requiring model retraining. The framework is evaluated through experiments using NLP metrics and a human study with five participants. Results show perceived differences in response style and structure across personalization conditions, with statistical analyses identifying learner attributes associated with measurable response changes. These findings provide preliminary evidence that prompt-based personalization can support adaptive behavior in LLM-powered educational agents.

\end{abstract}

\begin{links}
    \link{Code, Survey Links, and Dataset}{https://github.gatech.edu/sbasu7/IAAI27_JW_Personalization}
\end{links}


\section{Introduction}
\label{sec:introduction}
The aspiration to provide learners with educational experiences tailored to their individual needs is decades old. Bloom's 2 Sigma finding demonstrated that one-to-one tutoring can produce learning gains approximately two standard deviations above conventional classroom instruction, establishing personalization as a measurable educational objective~\cite{bloom1984}. Subsequent research has sought scalable approaches that approximate the benefits of individualized instruction through adaptive and intelligent learning systems~\cite{shute2012adaptive,aleven2017adaptive,bernacki2021review}.

The emergence of large language models (LLMs) has substantially expanded the potential for personalized learning. LLM-based teaching assistants can generate fluent, contextually relevant responses at scale and, when combined with retrieval-augmented generation (RAG), provide course-grounded instructional support across diverse disciplines~\cite{taneja2024jillchatgpt,kakar2024jillscaling,maiti2024jillinteraction}. However, their flexibility raises important questions regarding which learner characteristics should drive personalization, how personalization should be implemented, and whether such adaptations produce meaningfully different instructional interactions.

This paper addresses these questions through a prompt-engineering-based personalization framework for the Jill Watson virtual teaching assistant~\cite{goel2018jillwatson,taneja2024jillchatgpt,kakar2024jillscaling}. The framework operates at the level of individual student questions and combines learner preferences with question-level cognitive demand. Specifically, responses are personalized using six learner dimensions: metacognitive self-assessment, abstraction preference, verbosity preference, perceptual orientation, information processing style, and level of understanding. Cognitive demand is estimated using Bloom's Taxonomy, while learner preferences are based on the Felder-Silverman learning model~\cite{bloom1956taxonomy,felder1988learning}. Their combination yields 96 distinct learner profiles.

The framework is implemented entirely through structured prompt engineering over an existing RAG-based LLM tutor, enabling real-time personalization without model retraining or domain-specific authoring. We evaluate the approach using 2,910 generated responses spanning 30 student questions and 97 prompt configurations through NLP-based analyses, followed by a human evaluation with five participants. Results provide preliminary evidence that prompt-based personalization produces measurable and perceptible differences in response characteristics, supporting its potential for adaptive behavior in LLM-powered educational agents.

\section{Literature Review}
\label{sec:lit_review}

Bernacki et al.~\cite{bernacki2021review} proposed a broad personalization framework that characterize adaptive learning through four lenses: \emph{by whom, to what, how, and for what purpose}. Plass and Pawar~\cite{plass2020taxonomy} further distinguish adaptivity (system-driven) from adaptability (learner-driven), as well as macro- and micro-level adaptation. The present work focuses on micro-level cognitive adaptation that combines system-driven classification using Bloom's Taxonomy with learner-driven preference selection.

Earlier adaptive learning systems typically follow a diagnose-prescribe cycle involving learner modeling, action selection, and model updating~\cite{shute2012adaptive}. Learner models commonly represent prior knowledge and skill mastery, with adaptation primarily implemented through content selection or sequencing~\cite{xie2019systematic}. In contrast, this work shifts personalization toward response-form adaptation: answers remain grounded in a shared retrieved knowledge base while their abstraction, structure, verbosity, and cognitive framing are modified through prompt conditioning.

Recent LLM-based tutoring systems have expanded personalization through conversational interaction and retrieval-augmented generation (RAG), including extensions of Jill Watson~\cite{taneja2024jillchatgpt,kakar2024jillscaling,maiti2024jillinteraction}. Systems such as LPITutor~\cite{liu2025lpitutor}, PATS~\cite{li2025pats}, GPTutor~\cite{chen2024gptutor}, and AgentTutor~\cite{li2025agenttutor} adapt difficulty, personality, instructional content, or teaching workflows. Other approaches integrate LLMs with cognitive diagnosis models to improve learner modeling~\cite{dong2025kcd,liu2025lmcd,wei2025llm4cd,zhang2025llmcdm}. Persona and preference-aware systems, including Park et al. ~\cite{park2024conversation} and CloChat~\cite{ha2024clochat}, demonstrate the value of incorporating learner preferences into prompts.

Compared with these approaches, to enable  dynamic response adaptation at the individual interaction level, the proposed framework combines six learner dimensions with question-level cognitive analysis using Bloom's Taxonomy. It allows learners to actively specify response characteristics while automatically adapting cognitive depth. Thus, rather than primarily adapting content or learning pathways, the LLM/RAG based AI teaching assistant (Jill Watson) personalizes how shared instructional content is presented through prompt engineering. This hybrid integration of learner-driven adaptability and system-driven cognitive assessment represents an underexplored direction in LLM-based educational personalization.

\section{Personalization Framework Design}
\label{sec:personal_approach}

The proposed personalization framework consists of three components: learner characteristics, learner preferences, and a prompt-engineering mechanism that conditions the underlying large language model (LLM). Following the profile-conditioned approach of Park et al.~\cite{park2024conversation}, learner characteristics are represented through self-assessed metacognitive knowledge and the cognitive complexity of individual questions determined using Bloom's Taxonomy~\cite{bloom1956taxonomy}. This enables question-level personalization based on both learner understanding and query complexity.

Learner preferences are derived from the Felder-Silverman learning model~\cite{felder1988learning} and are treated as user-selected preferences rather than fixed psychometric classifications. Five dimensions are incorporated: abstraction, verbosity, perception, processing, and understanding. These dimensions control the granularity, length, communication orientation, engagement style, and organizational structure of generated responses, respectively. The corresponding categories are summarized in Table~\ref{tab0}.

\begin{table}[t]
\centering
\renewcommand{\arraystretch}{1.2}

\begin{tabular}{
|p{0.28\columnwidth}|
p{0.18\columnwidth}|
p{0.22\columnwidth}|
p{0.10\columnwidth}|
}
\hline
\textbf{Factors} & \textbf{Level I} & \textbf{Level II} & \textbf{Level III} \\ \hline

Metacognition (Self Assessment) & Beginner & Intermediate & Expert \\ \hline
Abstraction & In-Depth (Technical) & Big Picture (High-Level) & -- \\ \hline
Verbosity & Verbose & Concise & -- \\ \hline
Perception & Sensory & Intuitive & -- \\ \hline
Processing & Active & Reflective & -- \\ \hline
Understanding & Sequential & Global & -- \\ \hline

\end{tabular}

\caption{Learning Preferences and Levels}
\label{tab0}
\end{table}

The combination of learner characteristics and preferences produces 96 distinct learner profiles. In addition, each student question is automatically classified according to Bloom's Taxonomy, enabling dynamic adaptation at the individual interaction level. Learner preferences are explicitly selected by students, whereas cognitive demand is inferred by the system. This creates a hybrid framework combining learner-driven adaptability with system-driven adaptivity.

All learner attributes are encoded in an engineered prompt that conditions response generation. The prompt is integrated with Jill Watson's retrieval-augmented generation (RAG) pipeline and course-specific knowledge base, allowing personalization to modify the form and presentation of responses while preserving grounding in retrieved instructional content. An example prompt is shown below:

\begin{quote}
\textit{
I have a \textbf{beginner} level of knowledge in this topic. The Bloom’s Taxonomy category of my question is \textbf{Synthesis}. Please provide a \textbf{technical} and \textbf{concise} response, using a \textbf{sensory} communication style.I process information in a \textbf{reflective} way and prefer to understand concepts in a \textbf{global} manner. \\
Personalize the response based on my understanding and preferences listed in this prompt. The query is as follows: what approach should I take to best solve the Sheep and Wolves problem?
}
\end{quote}

The modular design allows learner preferences to be updated through the Jill Watson interface and incorporated into prompts at runtime, enabling scalable personalization without modifying or retraining the underlying LLM ~\cite{kakar2024jillscaling}.

\section{Research Questions}
\label{sec:research_question}

The objective of this study is to determine whether the proposed personalization framework produces distinct and perceptible response characteristics. Accordingly, we investigate the following research questions (RQs):

\begin{enumerate}

\item RQ1: To what extent are learner profiles associated with differences in the linguistic characteristics of LLM-generated responses?

\item RQ2: Which learner dimensions are most strongly associated with variations in response characteristics?

\item RQ3: Are the observed response characteristics consistent with the intended effects of the corresponding learner dimensions?

\end{enumerate}

RQ1 examines whether different learner profiles produce systematically different responses. RQ2 evaluates the relative contribution of individual learner dimensions to response characteristics, including semantic similarity, complexity, verbosity, abstraction, and processing style. RQ3 assesses whether observed differences align with the intended effects of each dimension; for example, whether higher verbosity produces longer responses and higher abstraction produces more technically complex explanations.

To address these questions, we combine automated NLP analyses with human evaluation. Descriptive and inferential statistical methods, including mixed-effects models, are used to quantify differences and associations across learner profiles and dimensions. 

\section{Experimental Design and NLP Evaluation}
\label{sec:synthetic_nlp}

Automated NLP analyses were conducted to determine whether personalization dimensions produce measurable differences in LLM-generated responses.

Thirty real-world student questions were selected from the Spring 2023 CS 7637 Knowledge-Based AI (KBAI) course at the Georgia Institute of Technology, covering all six Bloom's Taxonomy categories. Following Maiti and Goel~\cite{maiti2025ai_partner}, questions were classified using a fine-tuned BERT-based classifier trained on combined labeled datasets~\cite{gani2023exam,yahya2011blooms}. The classifier used \textit{bert-base-uncased} and achieved 0.92 test accuracy, with F1 scores of 0.88 - 0.94 across categories.

The six learner dimensions and their levels (Table~\ref{tab0}) produced 96 unique learner profiles. For each of the 30 questions, 97 prompts were generated: 96 personalized configurations and one non-personalized baseline, resulting in 2,910 responses. The prompt template and model configuration were held constant, with only learner-profile attributes varied. Responses were generated using GPT-4.1 with temperature set to 0 to minimize stochastic variation.

Responses were evaluated using four NLP dimensions: lexical similarity, semantic similarity, linguistic complexity, and verbosity. Semantic similarity was measured using 384-dimensional embeddings from \texttt{all-MiniLM-L6-v2}~\cite{wang2020minilm,hf_all_minilm_l6_v2}, followed by pairwise cosine similarity. Lexical overlap was measured using ROUGE, linguistic complexity using grade-level scores from \texttt{textstat}, and verbosity using \texttt{lexicon\_count}.

Descriptive analyses and ordinary least squares (OLS) regression were used to examine associations between learner dimensions and response characteristics. Together with the subsequent human evaluation, these analyses address the three research questions.

\section{Human Evaluation Study Design}
\label{sec:focus_group_design}

A human evaluation study involving five evaluators complemented the automated NLP analysis by assessing response characteristics that are difficult to capture automatically, including perceived abstraction, depth of understanding, and information-processing style. The evaluators were recruited from current and former students of the KBAI course at the Georgia Institute of Technology based on a pre-recruitment survey. The survey collected information about the evaluators’ subject-matter understanding, self-assessed expertise, and learning preferences.

One representative question was selected from each Bloom's Taxonomy category. The evaluation examined three personalization dimensions: self-assessment, abstraction, and processing style across 13 student profiles, including a non-personalized baseline. Evaluators were recruited through a screening survey capturing educational background, perceived competency, and learning preferences, and all data were anonymized.

Using Qualtrics, each evaluator assessed responses for all 13 profiles across the six questions, yielding 390 evaluations. Responses were rated on four dimensions: overall quality (0 - 10 scale), perceived complexity (novice to expert), abstraction level (non-technical, neutral, technical), and processing style (reflective, neutral, active).

For inferential analysis, the baseline was excluded, leaving 12 personalized profiles. Mixed-effects models were used with personalization dimensions as fixed effects and evaluator identity as a random effect to account for inter-rater variability.

Given the small sample of five evaluators, results are interpreted as exploratory rather than population-level evidence. The factorial design supports estimation of main effects but not interaction effects because of limited statistical power. Larger and more diverse samples are needed to establish generalization and examine interactions among personalization dimensions.

\section{Results and Discussion}

In this section, we present and briefly discuss the results from the NLP experiments and human evaluation study.

\subsection{NLP Experiments Results}
\label{sec:syn_nlp_results}

To qualitatively illustrate response differentiation, two responses generated for the question \textit{``What approach should I take to best solve the Sheep and Wolves problem?''} are compared. The first corresponds to a beginner profile preferring concise, non-technical, sensory, active, and sequential explanations. The second corresponds to an advanced profile preferring concise, technical, intuitive, reflective, and global explanations. Snippets from the responses are reproduced below as representative quotes.

\begin{quote}
\textit{Beginner, sensory, active, sequential profile:} ``Let's break it down into simple steps. Visualize the scenario, identify the rules, plan your moves, test different strategies.''
\end{quote}

\begin{quote}
\textit{Advanced, intuitive, reflective, global profile:} ``Define the problem space and constraints, identify primitive actions, map state transitions, apply explanation-based learning, and evaluate the solution globally.''
\end{quote}

The responses illustrate qualitative differences in instructional strategy: the beginner response emphasizes visualization, sequential steps, and experiential exploration, whereas the advanced response emphasizes formal decomposition, state-space reasoning, and abstraction. This provides qualitative evidence of response differentiation under different personalization configurations (RQ1).

\begin{figure}[t]
\centering
\includegraphics[width=0.75\columnwidth]{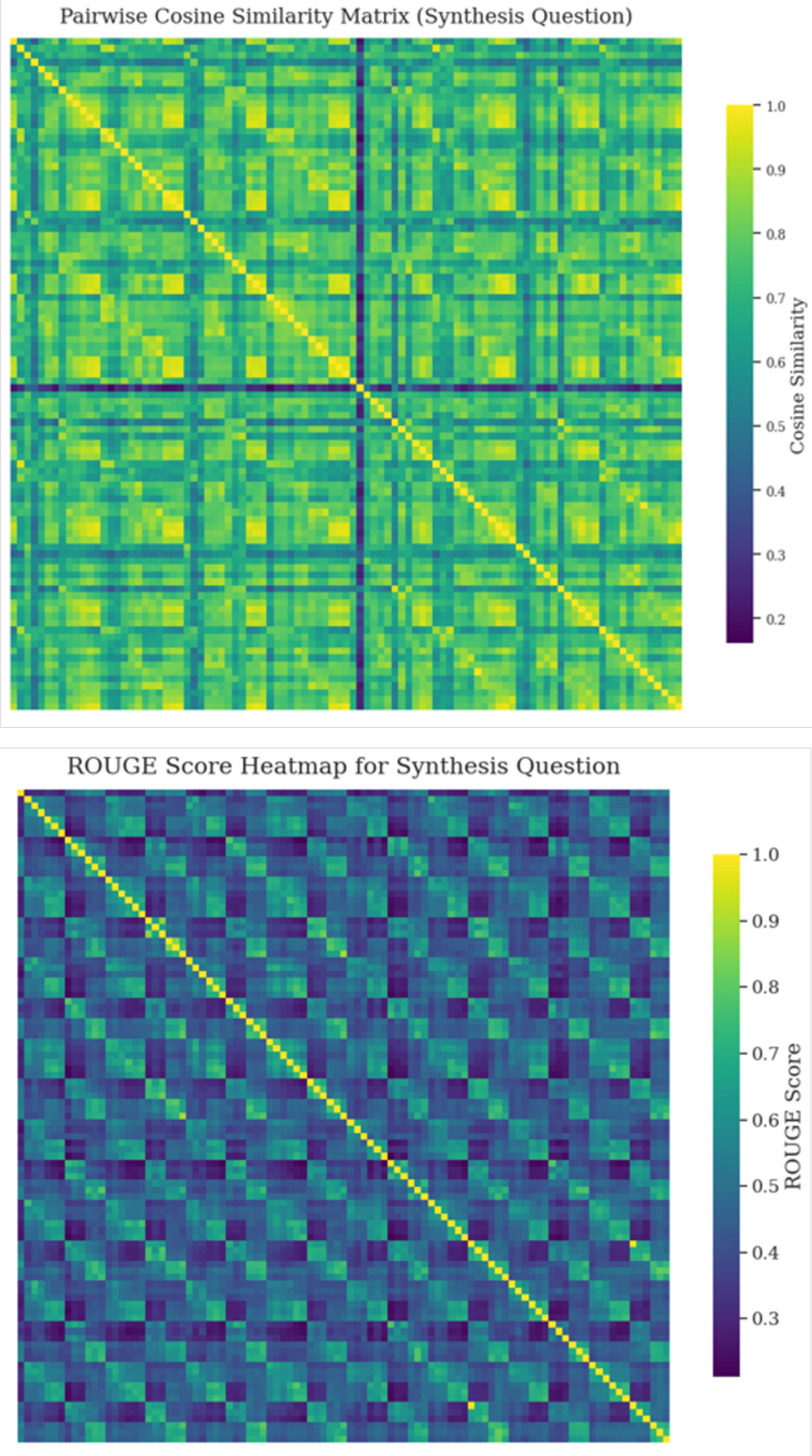}
\caption{Cosine similarity and ROUGE scores for 97 responses to a representative question.}
\label{fig_cos_rouge}
\end{figure}

To quantify response variation, all 97 responses for each question were encoded using SentenceTransformer \texttt{all-MiniLM-L6-v2}~\cite{wang2020minilm,hf_all_minilm_l6_v2}. Pairwise cosine similarity and ROUGE scores were computed. Figure~\ref{fig_cos_rouge} shows high semantic similarity but substantially lower lexical similarity, indicating that responses remain semantically grounded while varying in surface-level expression and structure.

\begin{figure}
\centering
\includegraphics[width=0.9\columnwidth]{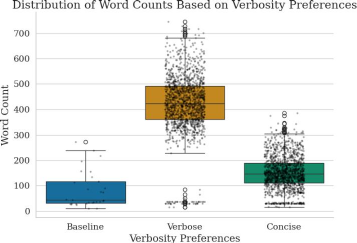}
\caption{Response word count across verbosity preference categories.}
\label{fig_verbosity_count}
\end{figure}

Figure~\ref{fig_verbosity_count} shows systematic variation in response length across verbosity preferences, providing evidence that the corresponding personalization dimension influences output length (RQ3).

\begin{figure}
\includegraphics[width=0.9\columnwidth]{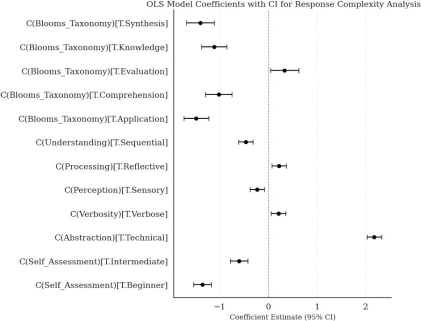}
\caption{OLS coefficient estimates for response complexity.}
\label{fig_ols_coeff}
\end{figure}

Complexity was measured using the \textit{textstat} grade-level score and analyzed using OLS regression with learner-profile attributes and Bloom's Taxonomy categories as predictors. As shown in Figure~\ref{fig_ols_coeff}, higher self-assessment, verbosity, reflective processing, and technical abstraction are associated with greater response complexity. Evaluation and Analysis questions also tend to produce more complex responses than other Bloom levels. These results indicate systematic associations between personalization dimensions and response characteristics (RQ2-RQ3).

Overall, the NLP analyses demonstrate measurable variation in response expression, length, and complexity across personalization conditions.

\subsection{Human Evaluation Study Results}

\begin{figure*}[t]
\includegraphics[width=0.95\textwidth]{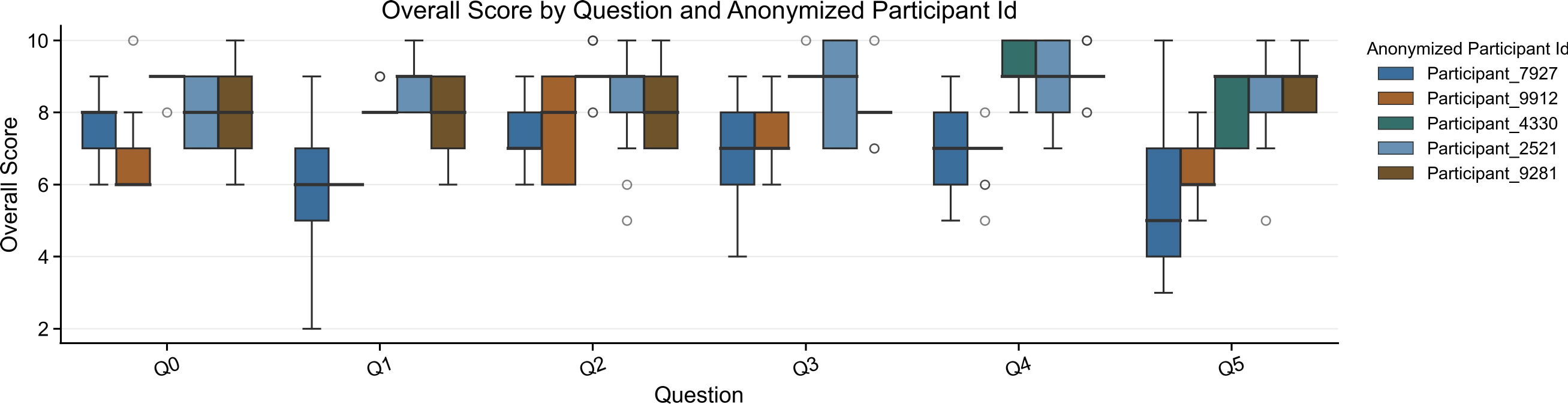}
\caption{Question-specific evaluation scores based on perceived accuracy and relevance.}
\label{fig_overall_score_fgs}
\end{figure*}

Evaluators rated the 13 responses for each question on overall accuracy and relevance. Figure~\ref{fig_overall_score_fgs} shows variation both across participants and across responses within participants, indicating that prompt-level personalization produced perceptible differences despite a shared RAG pipeline and knowledge base.

\begin{figure}
\includegraphics[width=0.9\columnwidth]{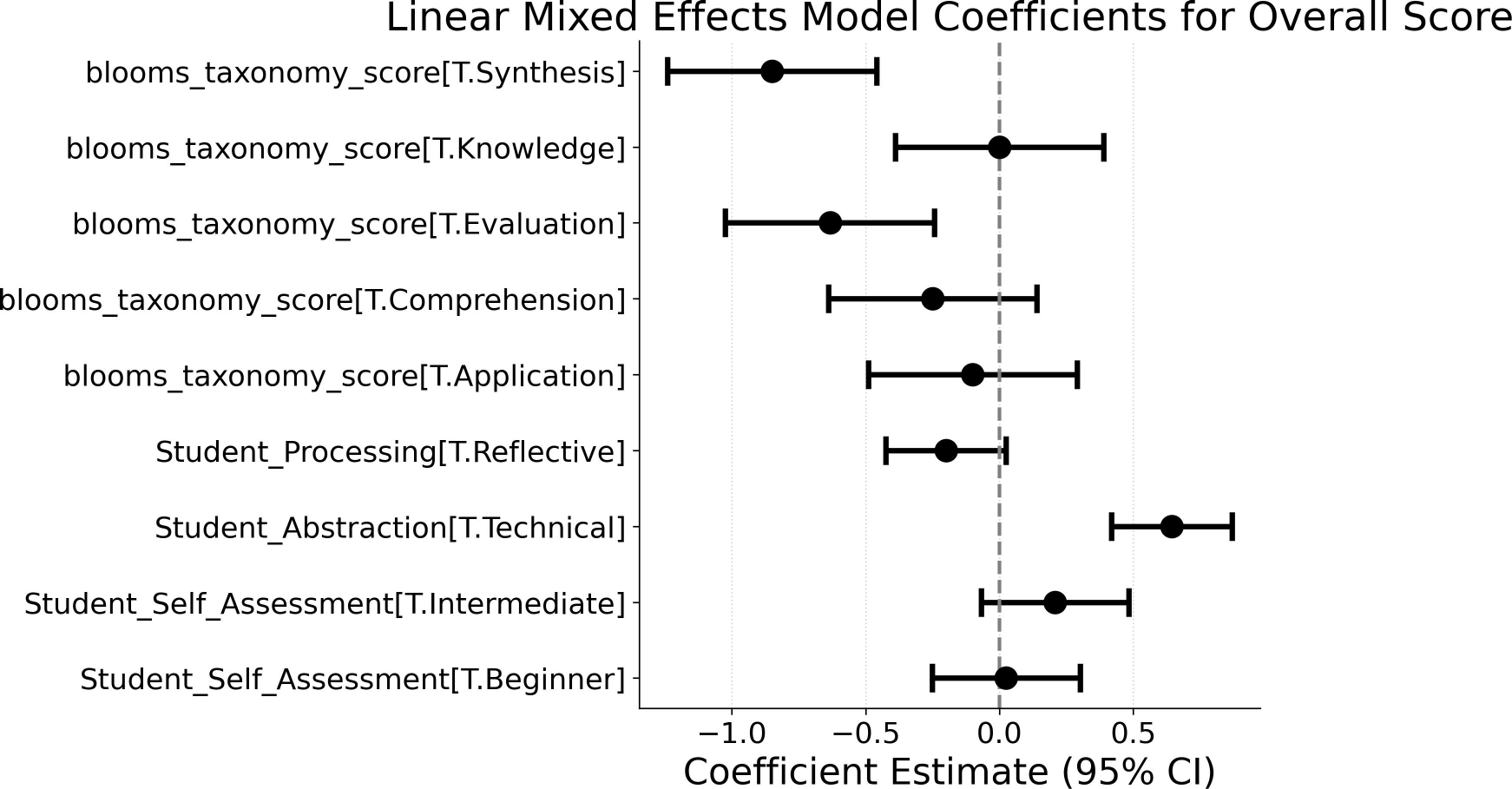}
\caption{Fixed-effect estimates for overall accuracy and relevance.}
\label{fig_ovr_score_forest}
\end{figure}

A linear mixed-effects model with personalization factors and Bloom's level as fixed effects and evaluator identity as a random effect showed that abstraction preference was associated with perceived response quality. Bloom's level was also associated with ratings, with more complex questions generally receiving lower scores (Fig.~\ref{fig_ovr_score_forest}).

\begin{figure}
\includegraphics[width=0.9\columnwidth]{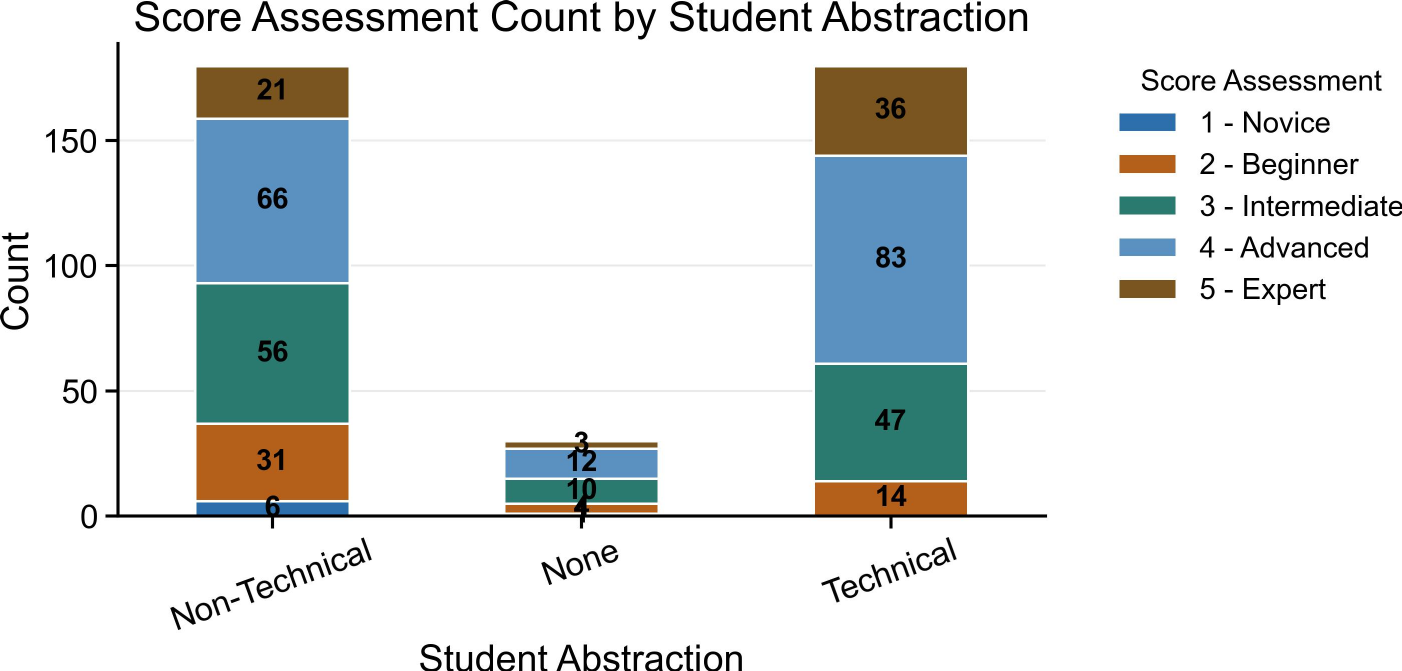}
\caption{Perceived response complexity by abstraction preference.}
\label{fig_assessment_levels_abstraction}
\end{figure}

\begin{figure}
\includegraphics[width=0.9\columnwidth]{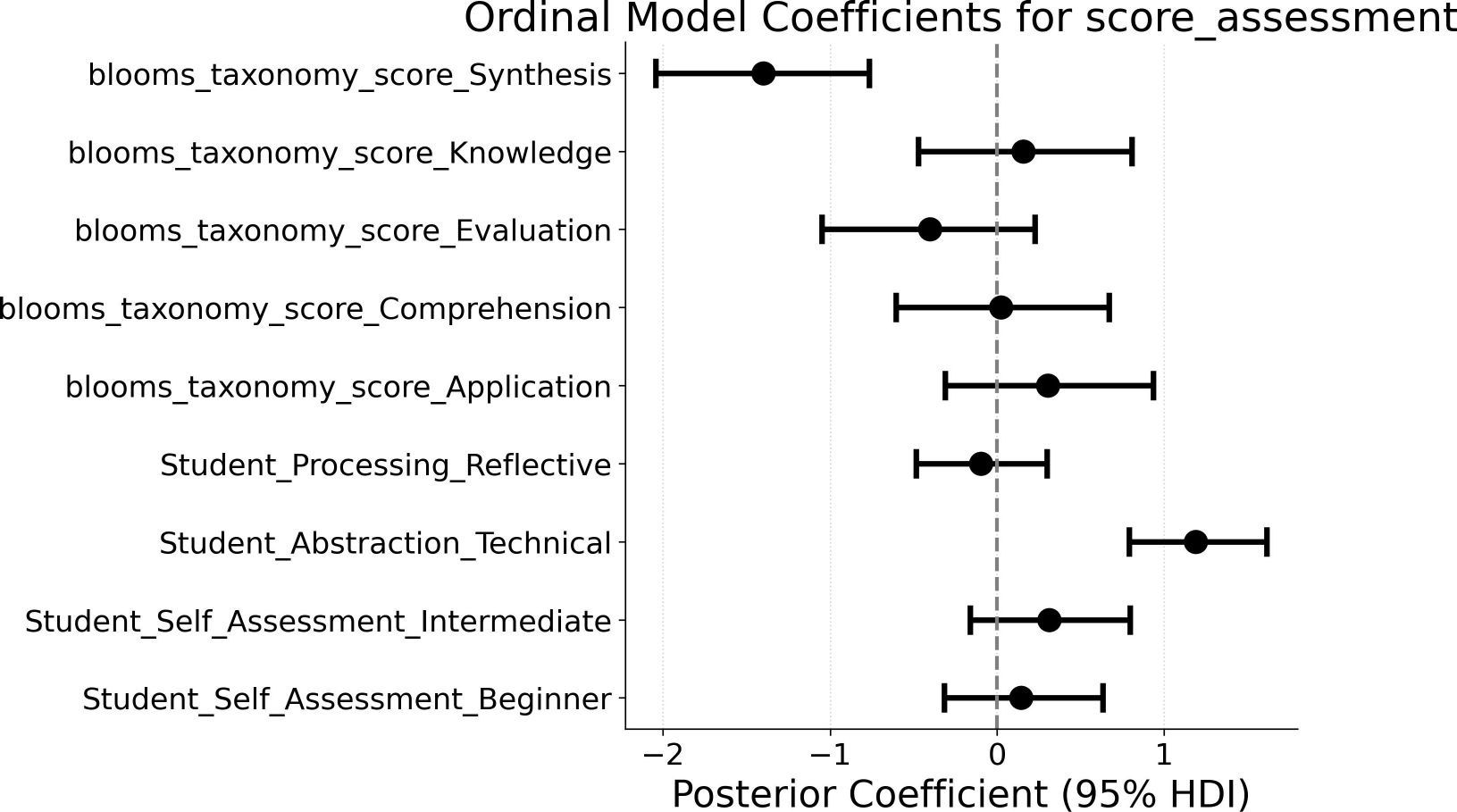}
\caption{Fixed-effect estimates for perceived response complexity.}
\label{fig_assessment_forest_plot}
\end{figure}

Evaluators also rated perceived response complexity on a five-level ordinal scale. Technical abstraction preferences were associated with higher perceived complexity (Fig.~\ref{fig_assessment_levels_abstraction}). A Bayesian ordinal mixed-effects model confirmed abstraction preference as a significant predictor of perceived complexity, while self-assessment was not significant (Fig.~\ref{fig_assessment_forest_plot}).

\begin{figure}
\includegraphics[width=0.9\columnwidth]{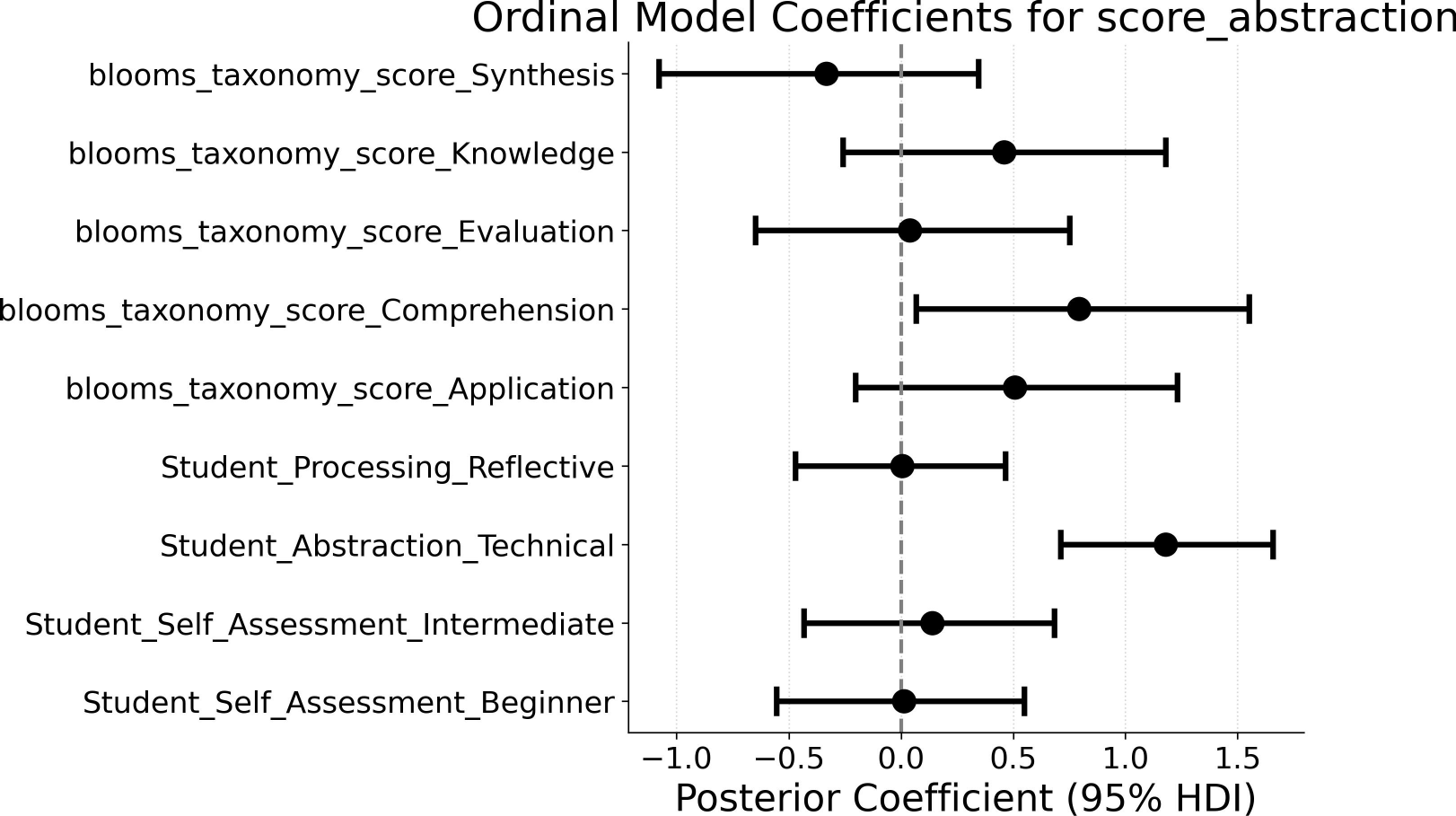}
\caption{Fixed-effect estimates for perceived abstraction level.}
\label{fig_abstraction_forest_plot}
\end{figure}

Perceived abstraction ratings generally aligned with the intended abstraction preferences. The corresponding Bayesian ordinal mixed-effects model identified both abstraction preference and Bloom's level as significant predictors (Fig.~\ref{fig_abstraction_forest_plot}), providing evidence that the intended abstraction differences were perceptible to evaluators.

\begin{figure}
\includegraphics[width=0.9\columnwidth]{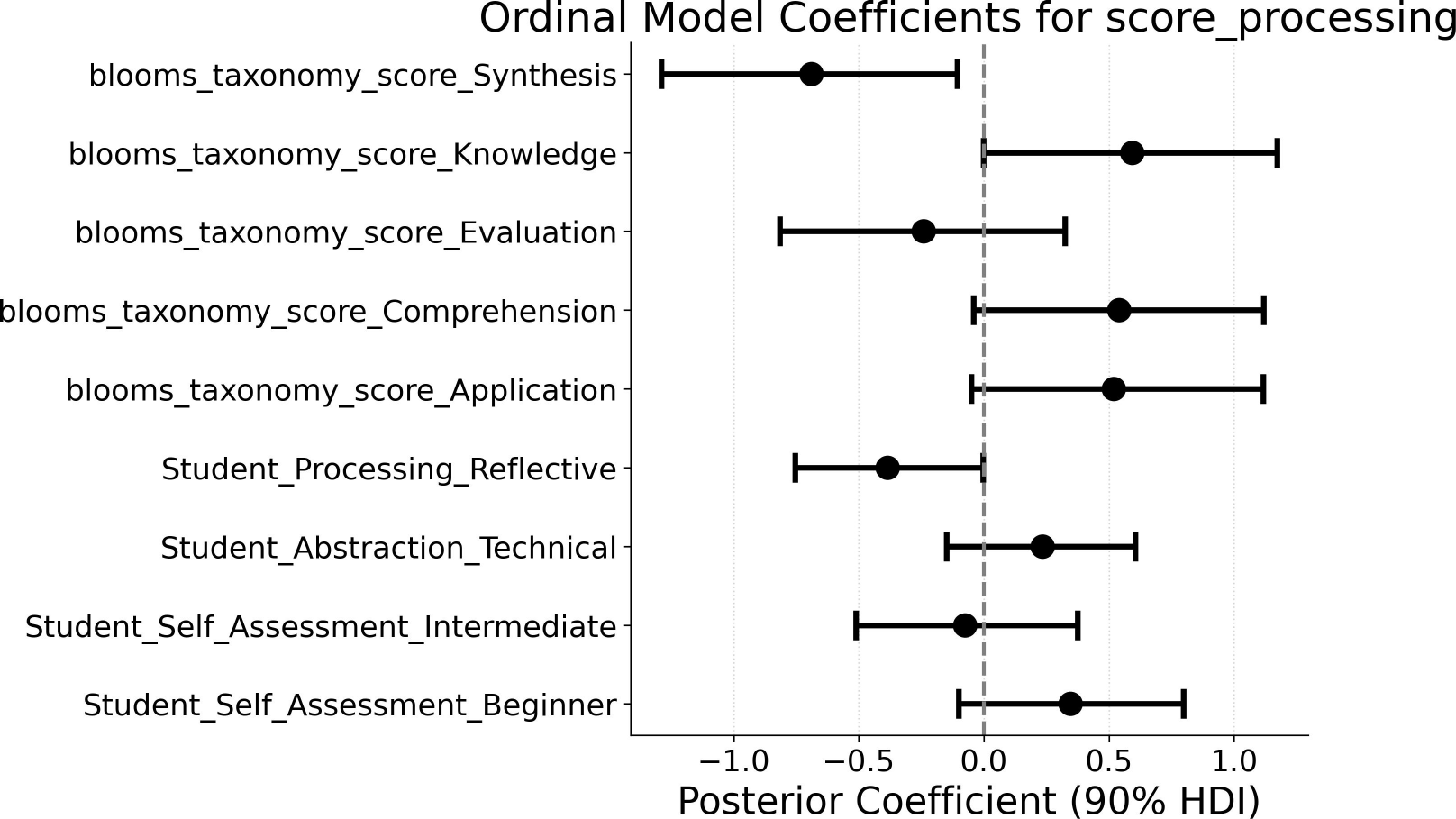}
\caption{Fixed-effect estimates for perceived processing style.}
\label{fig_processing_forest_plot}
\end{figure}

Similarly, processing preferences were reflected in evaluator ratings of response processing style. The ordinal mixed-effects model identified both processing preference and Bloom's level as significant predictors of perceived processing style (Fig.~\ref{fig_processing_forest_plot}).

Overall, the human evaluation provides evidence that prompt-based personalization produces perceptible differences in response quality, complexity, abstraction, and processing style. Abstraction and processing preferences were particularly consistent with their intended effects, while self-assessment showed weaker effects. These findings address RQ1-RQ3, although the small evaluator sample limits generalization. Larger studies are needed to assess robustness and interactions among personalization dimensions.

\section{Path to Deployment}

The proposed personalization module is designed for integration into the existing Jill Watson architecture that has already been deployed across multiple institutions, without modifying its core infrastructure~\cite{taneja2024jillchatgpt,kakar2024jillscaling,maiti2024jillinteraction}. Development and integration are targeted for completion by Spring 2027, followed by a pilot deployment in selected Georgia Institute of Technology courses in Summer 2027. The pilot will evaluate real-world performance and user feedback, informing subsequent refinement and broader deployment across additional courses and institutions in Fall 2027 - Spring 2028.

\section{Conclusions}

This paper presents a personalization framework for RAG and LLM based AI teaching assistants that enables flexible, scalable, modular, and real-time customization. The proposed framework emphasizes a hybrid approach between adaptability and adaptivity, enabling micro-level customization at the interaction level. Key contributions include:

\begin{enumerate}

    \item An engineered prompt that incorporates student cognitive ability, question complexity (Bloom’s Taxonomy), and learning preferences, generating 96 unique response configurations for question-level (micro) personalization.  
    
    \item Real-time adaptation of prompts based on learner-selected preferences, supported by system-level cognitive assessment using Bloom’s Taxonomy and a fine-tuned BERT-based classifier at each interaction.  
    
    \item A modular design that enables flexible integration of additional features, parameters, and prompt structures within the LLM/RAG (Jill Watson) architecture.  
    
    \item Scalability to support diverse learner models, knowledge bases, question banks, courses, and institutional settings without requiring domain-specific adaptation.
    
    \item This study addressed three research questions related to whether personalization leads to measurable response variation, which learner factors drive response differentiation, and whether learner preferences align with intended response characteristics. NLP-based experiments and human evaluation studies showed systematic response variation across conditions and identified key student profile parameters associated with changes in LLM-generated responses.  
    
\end{enumerate}

The current work primarily focuses on the proposed framework’s ability to personalize general-purpose LLM/RAG based AI teaching assistant's responses to individual student questions. Future work includes deployment of a UI-integrated personalized AI teaching assistant for large-scale classroom evaluation, A/B testing, and assessment of impacts on learning outcomes and student engagement.

\subsubsection{Acknowledgments} 
This research has been supported by NSF Grants 2112532 and 2247790 to the National AI Institute for Adult Learning and Online Education headquartered at Georgia Institute of Technology, Atlanta.

\bigskip

\bibliography{aaai2027}

@inproceedings{ha2024clochat,
  title        = {CloChat: Understanding how people customize, interact, and experience personas in large language models},
  author       = {Ha, Joon and Jeon, Hyeonjin and Han, Donghun and Seo, Joon and Oh, Changwoo},
  booktitle    = {Proceedings of the 2024 CHI Conference on Human Factors in Computing Systems},
  pages        = {1--24},
  year         = {2024},
  month        = may,
  publisher    = {ACM}
}

@book{bloom1956taxonomy,
  title     = {Taxonomy of Educational Objectives: The Classification of Educational Goals. Handbook I: Cognitive Domain},
  author    = {Bloom, Benjamin S. and Engelhart, Max D. and Furst, Edward J. and Hill, Walker H. and Krathwohl, David R.},
  year      = {1956},
  publisher = {Longman},
  address   = {New York}
}

@article{felder1988learning,
  title   = {Learning and Teaching Styles in Engineering Education},
  author  = {Felder, Richard M. and Silverman, Linda K.},
  journal = {Engineering Education},
  year    = {1988},
  volume  = {78},
  number  = {7},
  pages   = {674--681}
}

@article{plass2020taxonomy,
  author  = {Plass, Jan L. and Pawar, Shashank},
  title   = {Toward a Taxonomy of Adaptivity for Learning},
  journal = {Journal of Research on Technology in Education},
  volume  = {52},
  number  = {3},
  pages   = {275--300},
  year    = {2020},
  doi     = {10.1080/15391523.2020.1719943}
}

@article{bernacki2021review,
  author  = {Bernacki, Matthew L. and Greene, Meghan J. and
             Lobczowski, Nikki G.},
  title   = {A Systematic Review of Research on Personalized Learning:
             Personalized by Whom, to What, How, and for What Purpose(s)?},
  journal = {Educational Psychology Review},
  volume  = {33},
  number  = {4},
  pages   = {1675--1715},
  year    = {2021},
  doi     = {10.1007/s10648-021-09615-8}
}

@article{xie2019systematic,
  author  = {Xie, Haoran and Chu, Hui-Chun and Hwang, Gwo-Jen and
             Wang, Chun-Chieh},
  title   = {Trends and Development in Technology-Enhanced
             Adaptive/Personalized Learning: A Systematic Review of
             Journal Publications from 2007 to 2017},
  journal = {Computers \& Education},
  volume  = {140},
  pages   = {103599},
  year    = {2019},
  doi     = {10.1016/j.compedu.2019.103599}
}

@incollection{shute2012adaptive,
  author    = {Shute, Valerie J. and Zapata-Rivera, Diego},
  title     = {Adaptive Educational Systems},
  booktitle = {Adaptive Technologies for Training and Education},
  editor    = {Durlach, Paula J. and Lesgold, Alan M.},
  pages     = {7--27},
  publisher = {Cambridge University Press},
  address   = {Cambridge, MA},
  year      = {2012}
}

@incollection{aleven2017adaptive,
  author    = {Aleven, Vincent and McLaughlin, Elizabeth A. and
               Glenn, R. Amos and Koedinger, Kenneth R.},
  title     = {Instruction Based on Adaptive Learning Technologies},
  booktitle = {Handbook of Research on Learning and Instruction},
  edition   = {2nd},
  editor    = {Mayer, Richard E. and Alexander, Patricia A.},
  pages     = {522--560},
  publisher = {Routledge},
  address   = {New York, NY},
  year      = {2017}
}

@article{bloom1984,
  author  = {Bloom, Benjamin S.},
  title   = {The 2 Sigma Problem: The Search for Methods of Group Instruction
             as Effective as One-to-One Tutoring},
  journal = {Educational Researcher},
  volume  = {13},
  number  = {6},
  pages   = {4--16},
  year    = {1984}
}

@misc{yahya2011blooms,
  author       = {Yahya, A.},
  title        = {Bloom's Taxonomy Cognitive Levels Data Set},
  year         = {2011},
  howpublished = {Data set available at ResearchGate},
  doi          = {10.13140/RG.2.1.4932.3123},
  url          = {https://doi.org/10.13140/RG.2.1.4932.3123},
  note         = {Dataset}
}

@misc{gani2023exam,
  author       = {Gani, M. O. and Sangodiah, A.},
  title        = {Exam Question Datasets},
  year         = {2023},
  howpublished = {Figshare},
  doi          = {10.6084/m9.figshare.22597957},
  url          = {https://figshare.com/articles/dataset/Exam\_Question\_Datasets/22597957},
}

@inproceedings{goel2018jillwatson,
  author    = {Goel, Ashok K. and Polepeddi, Lalith},
  title     = {Jill {Watson}: A Virtual Teaching Assistant for Online Education},
  booktitle = {Learning Engineering for Online Education: Theoretical Contexts
               and Design-Based Examples},
  publisher = {Routledge},
  year      = {2018}
}

@inproceedings{taneja2024jillchatgpt,
  author    = {Taneja, Karan and Maiti, Pratyusha and Kakar, Sandeep and
               Guruprasad, Pranav and Rao, Sanjeev and Goel, Ashok K.},
  title     = {Jill {Watson}: A Virtual Teaching Assistant Powered by {ChatGPT}},
  booktitle = {Artificial Intelligence in Education (AIED 2024)},
  publisher = {Springer},
  year      = {2024}
}

@inproceedings{kakar2024jillscaling,
  author    = {Kakar, Sandeep and Maiti, Pratyusha and Taneja, Karan and
               Nandula, Alekhya and Nguyen, Gina and Zhao, Aiden and
               Nandan, Vrinda and Goel, Ashok K.},
  title     = {Jill {Watson}: Scaling and Deploying an {AI} Conversational
               Agent in Online Classrooms},
  booktitle = {International Conference on Intelligent Tutoring Systems},
  pages     = {78--90},
  publisher = {Springer Nature Switzerland},
  year      = {2024}
}

@article{maiti2024jillinteraction,
  author  = {Maiti, Pratyusha and Goel, Ashok K.},
  title   = {How Do Students Interact with an {LLM}-Powered Virtual
             Teaching Assistant in Different Educational Settings?},
  journal = {arXiv preprint arXiv:2407.17429},
  year    = {2024}
}

@inproceedings{maiti2025ai_partner,
  author    = {Maiti, P. and Goel, A.},
  title     = {Can an AI Partner Empower Learners to Ask Critical Questions?},
  booktitle = {Proceedings of the 30th International Conference on Intelligent User Interfaces (IUI)},
  pages     = {314--324},
  year      = {2025},
  month     = mar
}

@article{liu2025lpitutor,
  author  = {Liu, Zhensheng and Agrawal, Prateek and Singhal, Saurabh and
             Madaan, Vishu and Kumar, Mohit and Verma, Pawan Kumar},
  title   = {{LPITutor}: An {LLM}-Based Personalized Intelligent Tutoring
             System Using {RAG} and Prompt Engineering},
  journal = {PeerJ Computer Science},
  volume  = {11},
  pages   = {e2991},
  year    = {2025},
  doi     = {10.7717/peerj-cs.2991}
}

@inproceedings{park2024conversation,
  author    = {Park, Minju and Kim, Sojung and Lee, Seunghyun and
               Kwon, Soonwoo and Kim, Kyuseok},
  title     = {Empowering Personalized Learning Through a Conversation-Based
               Tutoring System with Student Modeling},
  booktitle = {Extended Abstracts of the CHI Conference on Human Factors
               in Computing Systems},
  pages     = {1--10},
  year      = {2024}
}

@article{li2025pats,
  author  = {Li, Mengxue and others},
  title   = {{PATS}: Personality-Aware Teaching Strategies with Large
             Language Model Tutors},
  journal = {arXiv preprint},
  year    = {2025}
}

@inproceedings{chen2024gptutor,
  author    = {Chen, Eason and Huang, Ray and Chen, Han-Shin and Tseng,
               Yuen-Hsien and Li, Liang-Yi},
  title     = {{GPTutor}: Great Personalized Tutor with Large Language Models
               for Personalized Learning Content Generation},
  booktitle = {Companion Proceedings of the ACM Web Conference},
  year      = {2024}
}

@inproceedings{li2025agenttutor,
  author    = {Li, Cheng and others},
  title     = {{AgentTutor}: Empowering Personalized Learning with Multi-Turn
               Interactive Teaching in Intelligent Education Systems},
  booktitle = {Proceedings of the AAAI Conference on Artificial Intelligence},
  year      = {2025}
}

@article{dong2025kcd,
  author  = {Dong, Zhiang and Chen, Jingyuan and Wu, Fei},
  title   = {Knowledge is Power: Harnessing Large Language Models for
             Enhanced Cognitive Diagnosis},
  journal = {arXiv preprint arXiv:2502.05556},
  year    = {2025}
}

@article{liu2025lmcd,
  author  = {Liu, Yu and others},
  title   = {{LMCD}: Language Models are Zero-Shot Cognitive Diagnosis
             Learners},
  journal = {arXiv preprint arXiv:2505.21239},
  year    = {2025}
}

@article{wei2025llm4cd,
  author  = {Wei, Guanhao and others},
  title   = {{LLM4CD}: Leveraging Large Language Models for Open-World
             Knowledge Augmented Cognitive Diagnosis},
  journal = {arXiv preprint arXiv:2505.13492},
  year    = {2025}
}

@article{zhang2025llmcdm,
  author  = {Zhang, Yibo and others},
  title   = {{LLM-CDM}: A Large Language Model Enhanced Cognitive Diagnosis
             for Intelligent Education},
  journal = {IEEE Transactions on Learning Technologies},
  year    = {2025}
}

@misc{hf_all_minilm_l6_v2,
  title        = {all-MiniLM-L6-v2 SentenceTransformer Model},
  author       = {{Sentence-Transformers Community on Hugging Face}},
  year         = {2024},
  howpublished = {\url{https://huggingface.co/sentence-transformers/all-MiniLM-L6-v2}},
  note         = {Accessed: 2026-06-26}
}

@article{wang2020minilm,
  title={MiniLM: Deep Self-Attention Distillation for Task-Agnostic Compression of Pre-Trained Transformers},
  author={Wang, Wenhui and others},
  journal={arXiv preprint arXiv:2002.10957},
  year={2020}
}


\end{document}